\documentclass[journal]{IEEEtran}
\IEEEoverridecommandlockouts
\usepackage{cite}
\usepackage{amsmath,amssymb,amsfonts}
\usepackage{algorithmic}
\usepackage{graphicx}
\usepackage{textcomp}
\usepackage{newtxtext,newtxmath}
\usepackage{xcolor}
\graphicspath{{figures/}}
\def\BibTeX{{\rm B\kern-.05em{\sc i\kern-.025em b}\kern-.08em
    T\kern-.1667em\lower.7ex\hbox{E}\kern-.125emX}}

\usepackage{booktabs}
\usepackage{pifont}
\usepackage{placeins}

\usepackage{multirow}
\begin{document}


\title{Constraint-Grounded Reinforcement Learning for Variable Impedance Control in Contact-Rich Robotic Insertion}

\author{Lin He and Min Deng$^{*}$%
\thanks{$^{*}$Corresponding author: Min Deng.}%
\thanks{The authors are with the Department of Civil and Environmental
Engineering, The University of Tennessee, Knoxville, TN 37996 USA
(e-mail: lynnhe@utk.edu; mindeng@utk.edu).}%
}

\maketitle

\begin{abstract}

In robotic insertion under uncertain contact, the axial force limit and the appropriate controller gain vary across tasks. As a result, a single fixed gain is unlikely to remain suitable across different task conditions, making conventional impedance controllers reliant on manual retuning. To eliminate manual retuning, we propose Constraint-Grounded Reinforcement Learning (CG-RL), a variable impedance framework for online gain adaptation. Conditioned on the force limit and contact feedback, the policy outputs a residual motion, an insertion rate, and a requested gain. The controller projects this gain into the admissible range without exposing the range itself to the policy. This separation allows a single policy to operate under different force limits without retraining or manual retuning. We evaluate CG-RL on simulated oblique insertion across five training seeds. CG-RL achieves an $85.8\pm7.7\%$ (mean $\pm$ SD) success rate of insertions without violating the force limit, while keeping the applied gain within the admissible range. As a comparison, a fixed-gain baseline using the midpoint gain achieves a success rate of $50.1\%$. The policy adapts its insertion rate continuously to the specified force limit and further generalizes to more permissive force limits above the training range. In contrast, the same actor without force-limit input does not exhibit this adaptation. The applied gain is guaranteed to remain within the admissible range, while force-limit satisfaction is validated empirically rather than guaranteed formally.
\end{abstract}


\section{Introduction}
\label{sec:introduction}

\begin{figure*}[t]
    \centering
    \includegraphics[width=\textwidth]
    {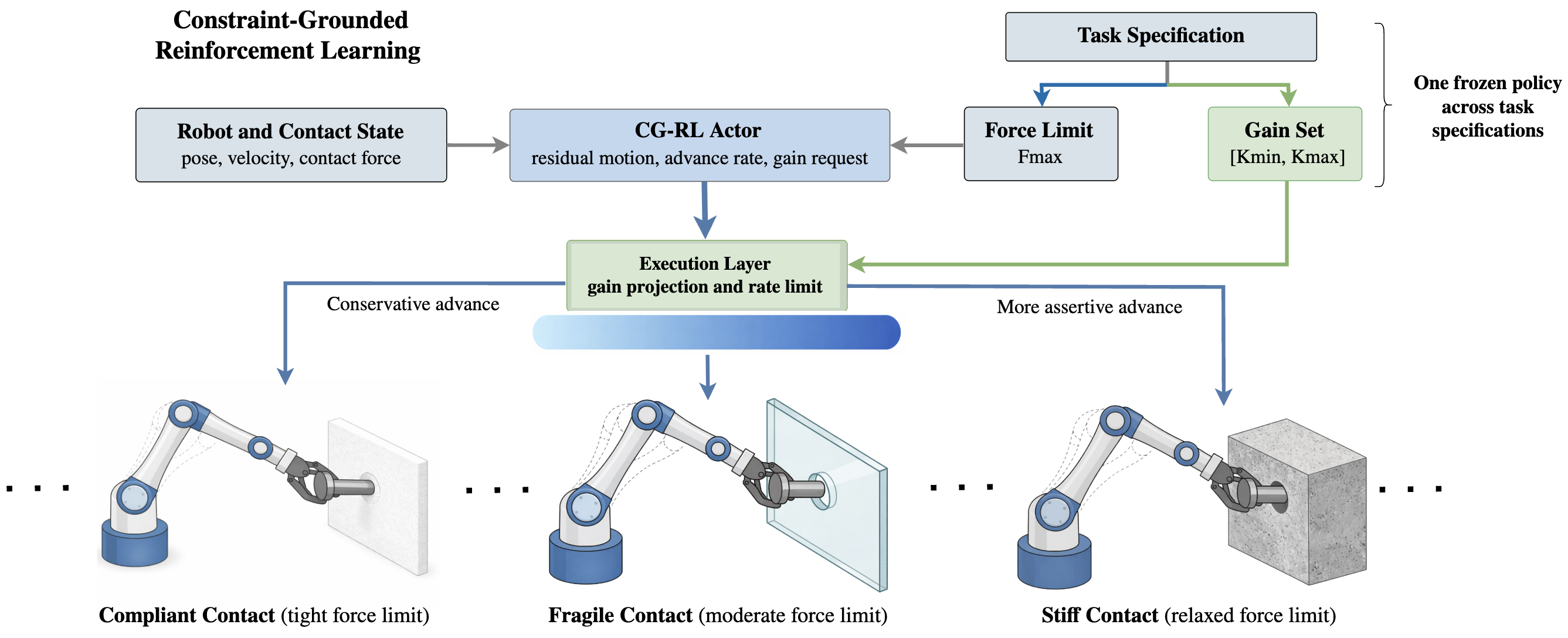}
    \caption{\textbf{Overview of CG-RL.}
    A single actor responds to the supplied force limit, while the execution layer independently enforces the admissible gain set across varying
    contact tasks. }
    \label{fig:overview}
\end{figure*}

Robotic manipulation is increasingly used in construction and manufacturing to address labor shortages, improve precision, and perform tasks in unsafe or ergonomically demanding settings~\cite{Davila2019ConstructionRobotics,deng2025llm_digital_twin,he2025llm, chen2026perception}. Yet contact-rich manipulation remains a central challenge, because robots must coordinate motion, force, and compliance in constrained or uncertain environments~\cite{Suomalainen2022ContactSurvey,Hogan1985Impedance}.
Insertion is a canonical contact-rich task underlying connector mating, press fitting, fastening, and component assembly~\cite{Whitney1982}. Oblique insertion is particularly relevant to load-bearing connections. In timber fastening, for example, inclined screws exploit the axial withdrawal resistance of the fastener and can carry greater loads than perpendicular screws~\cite{Bejtka2002InclinedScrews}. This mechanical advantage, however, comes with a more difficult insertion process. The inclined contact geometry generates substantial lateral forces and moments and makes the insertion more sensitive to misalignment and friction. As a result, the insertion becomes more prone to jamming and to damaging the inserted component or the surrounding workpiece~\cite{Apolinarska2021RLTimber}.

In practice, contact conditions also vary with material, thickness, support, clamping, and interface friction~\cite{Bock2015FutureConstruction}. Even for the same nominal material, these factors can change both the contact response and the load that the joint can tolerate~\cite{Yang2025DiffusionTimber,Wang2025CyclicGlulam}. 
As a result, a controller tuned for one task may apply excessive load in another. A conservative controller may instead move too slowly or fail to overcome the contact resistance. Selecting a controller gain and advance rate manually for every task is impractical, especially when interface friction is unknown before execution. We therefore seek a single policy that accepts the prescribed operating requirements and adjusts its motion, gain, and advance rate online. The policy should operate under fixture pose variation and unobserved contact variation without retraining or manual retuning.

Existing studies address separate aspects of this problem. Force-feedback and variable impedance controllers can regulate interaction forces and adjust compliance during execution. However, their force references, gain schedules, and adaptation rules are typically designed for a particular task or contact condition~\cite{AbuDakka2020VariableImpedance,Li2018ForceImpedanceTrajectory}. Learning-based variable impedance methods allow the robot to select motion and impedance online, but the available gain range is usually fixed when the controller is designed rather than specified for each task~\cite{Luo2019ICRA,MartinMartin2019VICES,BeltranHernandez2020ApplSci,Bogdanovic2020RAL}. Domain randomization improves robustness to unobserved variation in physical properties and controller parameters, but it treats these quantities as disturbances rather than as requirements supplied to the policy~\cite{Tobin2017DomainRand,andrychowicz2020learning}. Context-conditioned policies can adapt across tasks, yet the task variables or latent representations they rely on do not necessarily correspond to physical force or gain limits that the controller can enforce directly~\cite{rakelly2019efficient}. FORGE~\cite{noseworthy2025forge} is closest to our setting in that it conditions an assembly policy on a maximum allowable force; however, it treats the controller gains as randomized dynamics parameters rather than as actions selected by the policy within a task-specified gain range.

Existing methods therefore do not provide a single policy that selects its motion and gain online under an allowable force limit and an admissible gain set supplied independently for each task. Two difficulties arise in this setting. The first concerns enforcement: reward shaping alone cannot guarantee that the applied gain stays within the specified range, so the range is enforced at the controller instead. The second concerns attribution: once the controller projects and rate-limits the requested gain, the applied gain can change even when the raw policy request does not. A change in behavior can therefore be attributed to the policy only if the task specification, raw policy request, applied gain, and resulting contact response are recorded separately. These difficulties lead to the following research questions:
\begin{itemize}
    \item \textbf{RQ1:} How can a single learned policy select a suitable gain and advance rate online under a task-prescribed force limit, while the gain is kept within a task-prescribed admissible range, without retraining or retuning for each task?
    \item \textbf{RQ2:} Does such a policy change its insertion behavior with the prescribed force limit, beyond the behavior that controller enforcement alone produces?
\end{itemize}

To address these questions, we propose Constraint-Grounded Reinforcement Learning (CG-RL), a framework for learning variable impedance control policies for robotic insertion under explicitly specified contact requirements (Fig.~\ref{fig:overview}). Each task supplies an allowable axial force limit and an admissible gain set, which may vary independently, while interface friction remains hidden from the policy. \emph{Constraint grounding} refers to the explicit role assigned to each requirement. The force limit is observed by the actor and shapes its learning objective. The gain set, in contrast, is never observed by the actor; it is enforced by an execution layer that projects every requested gain into the admissible set before application. Gain-set membership is therefore guaranteed by construction, whereas compliance with the force limit is a learned property that we evaluate empirically rather than guarantee formally.

The main contributions of this paper are as follows.
\begin{enumerate}

    \item We formulate contact-rich robotic insertion as a constraint-grounded control problem in which an allowable axial force limit and an admissible controller gain set are specified independently, while interface friction remains unobserved. The formulation treats the permitted contact load and the available controller authority as distinct requirements with distinct enforcement mechanisms.
    
    \item We develop CG-RL, in which a single actor observes the force limit and interaction feedback and jointly selects a residual motion, an operational-space gain request, and an axial advance rate, while an execution layer projects and rate-limits the requested gain within the independently supplied gain set.

    \item We design an evaluation protocol that records the task specification, raw policy output, projected controller command, and physical response as separate signals, so that learned adaptation can be distinguished from controller enforcement. Using this protocol, we validate CG-RL across independent training seeds, paired interventions, dense force-limit sweeps, multiple gain sets, hidden friction conditions, and observation shifts, and identify the operating range within which the method remains effective.
\end{enumerate}

The remainder of this paper is organized as follows. Section~\ref{sec:related} reviews related work. Section~\ref{sec:problem_formulation} formulates the constraint-grounded insertion problem, and Section~\ref{sec:method} presents the CG-RL framework. Section~\ref{sec:experiments} reports the experimental evaluation. Section~\ref{sec:discussion} discusses the findings and limitations, and Section~\ref{sec:conclusion} concludes the paper.

\section{Related Work}
\label{sec:related}

\subsection{Compliant and Variable-Impedance Control}

Compliant control provides a structured basis for regulating physical interaction. Khatib~\cite{Khatib1987OperationalSpace} formulated manipulator dynamics directly in task space, unifying end-effector motion control and active force control in a single framework. Ott et al.~\cite{Ott2015HybridImpedanceAdmittance} subsequently developed a hybrid framework that spans impedance and admittance control, so that their complementary interaction characteristics can be combined within a common formulation. These works established the model-based foundation of physical interaction control. The resulting behavior, however, still depends on controller parameters that must be selected for each task.

Variable-impedance control relaxes this dependence by adjusting stiffness and damping during execution, but it raises the question of how to select suitable parameters as the interaction evolves~\cite{AbuDakka2020VariableImpedance,Zhou2025VICContactRich}. Learning-based approaches address this question using sensory feedback, demonstrations, or interaction experience. For example, Luo et al.~\cite{Luo2019ICRA} combined reinforcement learning with force/torque feedback and an operational-space controller and demonstrated high-precision assembly of a tight-fit gear system. Mart\'in-Mart\'in et al.~\cite{MartinMartin2019VICES} introduced variable impedance in end-effector space as an action representation for reinforcement learning, which improved sample efficiency, energy consumption, and transfer across contact-rich tasks. Other studies have learned force, trajectory, and impedance profiles for contact-sensitive manipulation, human-robot collaboration, and peg-in-hole assembly~\cite{Li2018ForceImpedanceTrajectory,Roveda2020,BeltranHernandez2020ApplSci,Bogdanovic2020RAL,Yang2022VariableImpedance,Kozlovsky2022PegInHoleImpedance}. More recently, Zhou et al.~\cite{Zhou2025VICContactRich} learned variable impedance from RGB-D and force/torque observations and demonstrated transfer to an industrial peg-in-hole task.

These results establish the value of learning impedance parameters online. In these methods, however, the learned gains remain within a range that is fixed when the controller is designed. The admissible gain set is not treated as a task-level requirement that can be specified before execution and enforced by the controller.

\subsection{Learning-Based Contact-Rich Manipulation}

Learning-based methods acquire contact behaviors from demonstrations or interaction data when contact dynamics are difficult to model analytically~\cite{ElgueaAguinaco2023RLContactRich}. Demonstration-based approaches have learned complex manipulation skills through behavioral cloning, implicit policy representations, and diffusion models~\cite{zhao2023learning,florence2022implicit,chi2023diffusion}. Diffusion Policy, for example, represented multimodal action distributions and improved performance across a diverse set of manipulation benchmarks~\cite{chi2023diffusion}. Other studies have incorporated tactile feedback, visual interaction cues, and multimodal object representations into policy learning~\cite{lee2019making,fazeli2019see,gao2022objectfolder}. In particular, Lee et al.~\cite{lee2019making} showed that a shared representation of vision and touch improved the sample efficiency of peg-in-hole policy learning and generalized across changes in geometry, configuration, and clearance. These results demonstrate that demonstrations and multimodal sensing can support complex contact behavior without a complete analytical model.

Reinforcement learning instead acquires contact strategies through interaction. Schoettler et al.~\cite{Schoettler2020MetaRL} trained a policy across simulated insertion tasks and adapted it to real tasks with fewer than 20 physical trials. Luo et al.~\cite{Luo2021RobustMultiModal} combined reinforcement learning, demonstrations, and multimodal observations, and reported greater speed and reliability than an engineered industrial baseline on an assembly benchmark. Policies whose actions include impedance or stiffness parameters further allow motion and compliance to be adjusted online during insertion~\cite{Yang2022VariableImpedance,Kozlovsky2022PegInHoleImpedance}. Together, these studies establish that learned policies can perform contact-rich insertion and assembly under substantial uncertainty.

In both families of methods, however, the primary objective is skill acquisition from data. These methods do not provide a deployment-time interface through which an allowable force limit or an admissible controller gain set can be specified and enforced. Moreover, when the allowable force and gain range can change independently, task success alone does not establish that the policy has responded to the active specification. It remains unclear whether explicit task requirements can command contact behavior predictably. This work addresses that question.

\subsection{Policy Learning under Task and Contact Variation}
\label{sec:constraint_policy}

Prior work on changing contact conditions differs mainly in how the variation is presented to the policy: as an explicit safety constraint, as task context, or as an unobserved disturbance. The first line of work, safe and constrained reinforcement learning, formulates requirements through cost limits, state or action constraints, safety filters, and control barrier functions~\cite{brunke2022safe,Wachi2024ConstraintSurvey,Cheng2019EndToEndSafeRL,Ames2019CBF,Fisac2019GeneralSafety}. Yao et al.~\cite{yao2023constraint} proposed constraint-conditioned policy optimization, which trains a single policy across cost thresholds and adapts to unseen thresholds without retraining. Sootla et al.~\cite{sootla2022saute} introduced Saut\'e RL, which augments the state with the remaining safety budget and targets almost-sure constraint satisfaction. More recently, Zhang et al.~\cite{Zhang2024SRLVIC} combined a safety critic and a recovery policy with online variable-stiffness control, and reported an improved trade-off between task completion and safety as well as transfer to a physical robot. These methods establish that safety thresholds and learned stiffness can be incorporated into policy learning. Their objective, however, is to satisfy a safety or risk criterion, and the admissible controller gain set is not treated as an independent requirement.

The second line of work, context-conditioned policies, adapts behavior using goals, task variables, or learned latent representations. Schaul et al.~\cite{Schaul2015UniversalValueFunctionApproximators} introduced universal value function approximators for goal-conditioned learning, and Rakelly et al.~\cite{rakelly2019efficient} learned latent task representations for rapid adaptation. Multi-task robotic learning has subsequently scaled policy training across larger task collections~\cite{Kalashnikov2021MTOpt}, and AutoMate~\cite{Tang2024AutoMate} demonstrated specialist and generalist assembly policies across diverse geometries with zero-shot transfer to physical systems. These results show that policy conditioning supports task generalization. The conditioning variables, however, usually identify a goal, task, or geometry rather than prescribe controller-level interaction limits. FORGE~\cite{noseworthy2025forge} provides a closer comparison by conditioning an assembly policy on a maximum allowable force. Its controller gains, however, are randomized as unobserved dynamics parameters rather than provided as task requirements or selected online by the policy.

The third line of work, domain randomization, addresses variation that is never revealed to the policy. It improves robustness by exposing policies to randomized physical parameters during training~\cite{Tobin2017DomainRand,peng2018sim,andrychowicz2020learning}. Factory and IndustReal applied this strategy to contact-rich industrial assembly by randomizing geometry, friction, dynamics, and sensing conditions~\cite{Narang2022Factory,Tang2023IndustReal}, and FORGE additionally randomized controller parameters to support robust force-aware assembly~\cite{noseworthy2025forge}. At deployment, these randomized quantities remain unobserved disturbances rather than requirements that specify how the task should be executed.

In summary, explicit conditioning and domain randomization serve complementary roles: the former communicates the requested operating condition, whereas the latter builds robustness to hidden physical variation. Nevertheless, none of the representative methods above jointly addresses independently specified force and gain requirements together with adaptation to unobserved contact variation. This gap motivates the problem formulation in Section~\ref{sec:problem_formulation}.

\section{Problem Formulation}
\label{sec:problem_formulation}

We consider oblique peg insertion along a nominal approach trajectory with variations in fixture pose and unobserved contact conditions. The force limit and admissible gain range are specified independently for each task. The actor observes pose perturbations through changes in the relative geometry, while variations in interface friction affect the contact dynamics but remain unobserved.

\subsection{Task Specification and Requirements}
\label{subsec:task_definition}

Each insertion task is associated with two independently specified requirements: an allowable axial force limit and an admissible controller gain set. We represent them by
\begin{equation}
c
=
\left(
F_{\max},
\mathcal{K}
\right),
\qquad
\mathcal{K}
=
\left[
K_{\min},
K_{\max}
\right],
\label{eq:descriptor}
\end{equation}
where $F_{\max}>0$ is the allowable axial force limit and $\mathcal{K}$ is the admissible interval for the scalar gain of the operational-space controller.

To define the physical quantity bounded by $F_{\max}$, let
$\hat{\mathbf d}$ denote the unit vector along the insertion axis in the world frame. Let $\mathbf f_t^{\mathrm{cont}}$ denote the net contact reaction at time step $t$. The magnitude of the axial contact reaction is
\begin{equation}
F_t^{\mathrm{ax}}
=
\left|
\hat{\mathbf d}^{\top}
\mathbf f_t^{\mathrm{cont}}
\right|.
\label{eq:axial_force}
\end{equation}
The force limit applies to this axial component rather than to the magnitude of the full contact force vector.

Let $\mathcal{F}\subset\mathbb{R}_{>0}$ denote the set of allowable force
limits, and let
\begin{equation}
\mathcal{K}_{\mathrm{sys}}
=
\left[
K_{\mathrm{sys}}^{\min},
K_{\mathrm{sys}}^{\max}
\right]
\subset\mathbb{R}_{>0}
\label{eq:system_gain_range}
\end{equation}
denote the system gain range of the controller. A gain set is admissible when $0<K_{\min}<K_{\max}$ and $\mathcal{K}\subseteq\mathcal{K}_{\mathrm{sys}}$. The task specification space is therefore
\begin{equation}
\mathcal{C}
=
\left\{
(F_{\max},\mathcal{K})
\;\middle|\;
F_{\max}\in\mathcal{F},\;
0<K_{\min}<K_{\max},\;
\mathcal{K}\subseteq\mathcal{K}_{\mathrm{sys}}
\right\}.
\label{eq:specification_space}
\end{equation}

During training, the force limit and gain set are sampled independently:
\begin{equation}
p_{\mathcal{C}}(c)
=
p_F(F_{\max})
p_{\mathcal{K}}(\mathcal{K}).
\label{eq:independent_specification}
\end{equation}
This factorization allows the same force limit to be paired with different gain sets, and vice versa.

The two components of $c$ play different roles. The force limit specifies the contact load permitted by the task. The gain set specifies the range of gains the controller is permitted to apply during execution, which we refer to as the controller authority. Although changing the scalar gain modulates the closed-loop stiffness, $K_{\min}$ and $K_{\max}$ are controller gain settings rather than estimates of material stiffness.

The specification is assumed to be available before execution, for example from task design requirements or prior characterization. Estimating the force limit or gain set online is outside the scope of this work.

An insertion is successful if the peg reaches a prescribed terminal set $\mathcal{X}_{\mathrm{goal}}$, representing the required insertion depth and pose accuracy, within a finite horizon $T$. Beyond completion, execution should be efficient and should limit exceedance of the axial force limit. We define the instantaneous force exceedance as
\begin{equation}
v_t^{F}
=
\left[
F_t^{\mathrm{ax}}
-
F_{\max}
\right]_{+},
\label{eq:force_violation}
\end{equation}
where $[z]_{+}=\max(z,0)$. The applied controller gain, by contrast, is
required to satisfy
\begin{equation}
K_t
\in
\mathcal{K},
\qquad
\forall t\in\{0,\ldots,T-1\}.
\label{eq:gain_requirement}
\end{equation}

The two requirements have different enforcement mechanisms. The gain requirement \eqref{eq:gain_requirement} can be enforced exactly at the controller interface. The force requirement is not enforced directly at the controller interface in this work, because the axial force is a closed-loop physical response that depends on pose error, contact state, friction, and the selected control actions. Satisfaction of the force limit is therefore learned and evaluated empirically rather than formally guaranteed.

\subsection{Decision Problem under Task Specifications}
\label{subsec:conditioned_control}

Let $x_t\in\mathcal{X}$ denote the underlying state of the robot and contact system, and let $z_t\in\mathcal{Z}$ denote the physical information available to the policy. This information includes relative geometry, robot motion, contact feedback, and controller state. The policy observation is
\begin{equation}
o_t
=
\left(
z_t,
\widetilde{F}_{\max}
\right),
\label{eq:policy_observation}
\end{equation}
where $\widetilde{F}_{\max}$ is the normalized force-limit coordinate. The gain set $\mathcal{K}$ is not part of the observation; it is supplied directly to the execution layer, which enforces set membership by projection.

We denote the unobserved contact condition by $\eta$. In this work, $\eta$ corresponds to interface friction and affects the transition dynamics:
\begin{equation}
x_{t+1}
\sim
\mathcal{P}_{\eta}
\left(
\cdot
\mid
x_t,u_t,K_t
\right),
\label{eq:hidden_dynamics}
\end{equation}
where $u_t$ is the command applied to the robot. The policy does not observe $\eta$ and can respond only to its effects on the observed motion and contact feedback.

The policy is given by
\begin{equation}
a_t
\sim
\pi_{\theta}
\left(
\cdot
\mid
o_t
\right).
\label{eq:conditioned_policy}
\end{equation}
Its normalized action is
\begin{equation}
a_t
=
\left[
\left(a_t^{\mathrm{res}}\right)^{\top},
a_t^{K},
a_t^{v}
\right]^{\top}
\in[-1,1]^8,
\label{eq:policy_action}
\end{equation}
where $a_t^{\mathrm{res}}\in[-1,1]^6$ is the residual pose action, $a_t^{K}\in[-1,1]$ is the scalar gain action, and $a_t^{v}\in[-1,1]$ is the advance-rate action.

Let $h_t$ denote the internal state of the execution layer, including the current primitive phase and the previously applied gain. An execution map converts the policy action and task specification into the command applied to
the robot:
\begin{equation}
\left(
u_t,
K_t,
h_{t+1}
\right)
=
\mathcal{G}
\left(
a_t,c,z_t,h_t
\right).
\label{eq:execution_map}
\end{equation}
The execution map is required to produce an applied gain that satisfies \eqref{eq:gain_requirement}. Note that $\mathcal{G}$ receives the complete task specification $c$, whereas the actor receives only $F_{\max}$; this is the formal counterpart of the constraint grounding described in Section~\ref{sec:introduction}. The formulation also keeps the raw policy action $a_t$, the applied command $(u_t,K_t)$, and the resulting physical response as separate quantities. This is necessary because controller enforcement can change $K_t$ even when $a_t^{K}$ is unchanged. This separation underlies the evaluation protocol in Section~\ref{sec:experiments}.

\subsection{Learning Objective}
\label{subsec:learning_objective}

At the beginning of each episode, a task specification
$c\sim p_{\mathcal{C}}$ and a hidden contact condition
$\eta\sim p_{\eta}$ are sampled independently and remain fixed throughout the
episode. Their joint distribution satisfies
\begin{equation}
p(c,\eta)
=
p_{\mathcal{C}}(c)
p_{\eta}(\eta).
\label{eq:specification_contact_independence}
\end{equation}

Let $\xi\sim p_{\theta}(\cdot\mid c,\eta)$ denote the trajectory induced by
the policy under specification $c$ and hidden condition $\eta$. The learning
objective is
\begin{equation}
\theta^{*}
=
\operatorname*{arg\,max}_{\theta}
\;
\mathbb{E}_{
\substack{
c\sim p_{\mathcal{C}},\,
\eta\sim p_{\eta},\\
\xi\sim p_{\theta}(\cdot\mid c,\eta)
}
}
\left[
\sum_{t=0}^{T-1}
\gamma^{t}
r(x_t,a_t,c)
\right],
\label{eq:learning_objective}
\end{equation}
where $\gamma\in(0,1]$ is the discount factor, subject to
\begin{equation}
K_t\in\mathcal{K},
\qquad
\forall t\in\{0,\ldots,T-1\}.
\label{eq:learning_gain_requirement}
\end{equation}

The reward captures insertion progress, task completion, execution efficiency, and exceedance of the allowable axial force limit; its complete definition is given in Section~\ref{sec:method}. The gain constraint \eqref{eq:learning_gain_requirement} is enforced by the execution map rather than through the reward.

The solution sought is therefore a single actor, coupled to a fixed execution map, that is optimized across force limits and hidden contact conditions. The actor should respond continuously to the prescribed $F_{\max}$ and, through interaction feedback, to the hidden contact condition $\eta$.

\section{Method}
\label{sec:method}

\subsection{Method Overview}
\label{subsec:method_overview}

Constraint-Grounded Reinforcement Learning (CG-RL) implements the formulation in Section~\ref{sec:problem_formulation} using four components: a learned actor, a geometric insertion primitive, a gain execution layer, and a variable-gain operational-space controller. At each policy step, the actor observes the robot and contact state $z_t$ together with the normalized force limit $\widetilde F_{\max}$. The actor outputs a bounded pose residual, an advance-rate action, and a gain request. The admissible gain set $\mathcal K$ is not included in the actor observation. Instead, the execution layer receives $\mathcal K$ directly and projects the requested gain into this set before application.

The force limit $F_{\max}$ also appears in the force-dependent reward terms. Interface friction is not observed directly and affects the actor only through interaction feedback. Section~\ref{subsec:policy_observation} describes the policy observation, and Section~\ref{subsec:primitive_actions} defines the nominal primitive and policy actions. Sections~\ref{subsec:gain_execution} and~\ref{subsec:reward_design} present gain execution and reward design, respectively. Section~\ref{subsec:training_deployment} describes training and deployment.

\subsection{Policy Observation}
\label{subsec:policy_observation}

The actor receives a 35-dimensional observation. The observation contains 28 kinematic and geometric quantities: the hand position relative to the fixture (3), a canonical hand quaternion (4), hand linear and angular velocity (6), a primitive state summary (8), the peg position relative to the hole (3), and a canonical peg quaternion (4). Seven additional quantities describe interaction and control: the axial contact reaction (1), a contact moment proxy (3), the binary contact state (1), the applied gain normalized by the system gain range $\mathcal{K}_{\mathrm{sys}}$ (1), and the normalized force limit (1).

The force limit is provided to the actor through the normalized scalar
\begin{equation}
\widetilde{F}_{\max}
=
\frac{F_{\max}}{\bar{F}},
\qquad
\bar{F}=9.0\,\mathrm{N}.
\label{eq:force_limit_normalization}
\end{equation}
Two quantities are deliberately excluded from the observation. The gain set endpoints $K_{\min}$ and $K_{\max}$ are supplied directly to the execution layer. The interface friction coefficient is varied independently of the task specification during training, and its effects reach the actor only through the observed motion and contact response. The actor can therefore respond to the realized interaction without observing or identifying the underlying friction value.

\subsection{Nominal Primitive and Policy Actions}
\label{subsec:primitive_actions}

The policy action is the eight-dimensional vector defined in \eqref{eq:policy_action}. Its first six coordinates, $a_t^{\mathrm{res}}$, specify translational and rotational pose residuals, $a_t^K$ is the raw normalized gain action, and $a_t^v$ is the raw normalized advance-rate action.

A CAD-defined geometric primitive supplies the nominal insertion sequence, which consists of five phases: alignment, precontact insertion, contact ramp and hold, recentering, and final insertion. The primitive is registered using the current fixture transform obtained directly from the simulator state. The method therefore assumes access to the ground-truth fixture pose and does not model fixture perception or pose estimation. During training, fixture pose perturbations are introduced progressively through a success-gated curriculum.

Let $\mathbf{T}_t^{\mathrm{prim}}$ denote the registered primitive pose and let $\Delta\mathbf{T}_t^{\mathrm{res}}(a_t^{\mathrm{res}})$ denote the bounded pose correction generated from the policy action. The desired pose is
\begin{equation}
\mathbf{T}_t^{d}
=
\mathbf{T}_t^{\mathrm{prim}}
\circ
\Delta\mathbf{T}_t^{\mathrm{res}}
\left(a_t^{\mathrm{res}}\right),
\label{eq:residual_pose_command}
\end{equation}
where $\circ$ denotes pose composition. Each translational residual coordinate is bounded by $\pm2$\,mm and each rotational (axis-angle) coordinate by $\pm3^\circ$. The execution layer scales these residuals by 0.25, which gives effective per-coordinate bounds of $\pm0.5$\,mm and $\pm0.75^\circ$, respectively. The residual is an absolute bounded offset around the primitive rather than an incremental command that accumulates over time.

During the contact ramp and hold phase, the learned pose residual is set to zero. This creates a standardized contact interval in which the effects of the gain and advance rate are not confounded by simultaneous policy-driven pose corrections.

The advance-rate action is mapped to a positive, dimensionless rate multiplier:
\begin{equation}
s_t^v
=
s_{\min}
+
\frac{\operatorname{clip}(a_t^v,-1,1)+1}{2}
\left(s_{\max}-s_{\min}\right),
\label{eq:advance_rate_mapping}
\end{equation}
where $0<s_{\min}<s_{\max}$. We use $s_{\min}=0.5$ and $s_{\max}=1.5$; the remaining fixed constants are summarized in Table~\ref{tab:implementation_constants}. During the precontact and final insertion phases, $s_t^v$ scales the nominal axial increment of $0.25$\,mm per policy step. The contact ramp follows a fixed probing motion so that contact responses are measured under a common excitation.

Note that the rate action controls how fast the desired insertion depth advances; it does not directly command force or end-effector velocity. The resulting motion and axial reaction depend jointly on this advance, the applied controller gain, the current pose error, and the contact dynamics.

\subsection{Gain Selection and Controller Execution}
\label{subsec:gain_execution}

The gain execution path maps the raw gain action to the applied gain in three stages: the requested gain, the projected gain, and the applied gain. Each stage is recorded as a separate signal. Let
\[
\mathcal K_{\mathrm{sys}}
=
[K_{\mathrm{sys}}^{\min},K_{\mathrm{sys}}^{\max}]
\]
denote the system gain range of the controller defined in \eqref{eq:system_gain_range}. The raw gain action is first mapped into this fixed range and then projected into the gain set specified for the current task:
\begin{align}
K_t^{\mathrm{req}}
&=
K_{\mathrm{sys}}^{\min}
+
\frac{
\operatorname{clip}(a_t^K,-1,1)+1
}{2}
\left(
K_{\mathrm{sys}}^{\max}
-
K_{\mathrm{sys}}^{\min}
\right),
\label{eq:requested_gain_mapping}
\\
K_t^{\mathrm{proj}}
&=
\Pi_{\mathcal K}
\left(
K_t^{\mathrm{req}}
\right)
=
\operatorname{clip}
\left(
K_t^{\mathrm{req}},
K_{\min},
K_{\max}
\right).
\label{eq:gain_projection}
\end{align}
Because \eqref{eq:requested_gain_mapping} is independent of $\mathcal K$, a given raw action corresponds to the same requested gain in every task. Consequently, changes in $a_t^K$ can be interpreted independently of the changes introduced by projection.

The projected gain is then passed through a first-order rate limiter. Let $\dot K_{\max}^{\mathrm{cmd}}$ denote the maximum gain-change rate and let $\Delta t$ be the policy interval. The applied gain is
\begin{equation}
\begin{aligned}
K_t
={}&
\Pi_{\mathcal K}
\Bigl[
K_{t-1}
+
\operatorname{clip}
\bigl(
K_t^{\mathrm{proj}}-K_{t-1},
\\
&\qquad\quad
-\dot K_{\max}^{\mathrm{cmd}}\Delta t,\;
\dot K_{\max}^{\mathrm{cmd}}\Delta t
\bigr)
\Bigr].
\end{aligned}
\label{eq:applied_gain}
\end{equation}
The final projection ensures that $K_t\in\mathcal K$ at every execution step, including immediately after a new task specification is assigned. The applied scalar gain is replicated across the six operational-space axes, yielding $\mathbf K_t=K_t\mathbf I_6$. Let $\mathbf x_t^d$ and $\dot{\mathbf x}_t^d$ denote the desired task pose and velocity, and let $\mathbf x_t$ and $\dot{\mathbf x}_t$ denote their measured counterparts. The controller first forms the task-space acceleration command
\begin{equation}
\ddot{\mathbf x}_t^{\mathrm{cmd}}
=
\mathbf K_t
\left(
\mathbf x_t^d-\mathbf x_t
\right)
+
\mathbf D_t
\left(
\dot{\mathbf x}_t^d-\dot{\mathbf x}_t
\right),
\label{eq:osc_motion_law}
\end{equation}
where $\mathbf D_t=2\zeta\mathbf K_t^{1/2}$ with $\zeta=1$. The desired velocity is zero within each low-level control interval, and motion is generated by updating the desired pose through the insertion primitive. The damping gain is recomputed from the applied proportional gain with a fixed damping ratio, so changing $K_t$ does not introduce an independently selected damping action.

With inertial decoupling, the commanded task-space wrench and joint torque are
\begin{equation}
\mathbf f_t^{\mathrm{cmd}}
=
\boldsymbol{\Lambda}_t
\ddot{\mathbf x}_t^{\mathrm{cmd}},
\qquad
\boldsymbol{\tau}_t
=
\mathbf J_t^{\top}
\mathbf f_t^{\mathrm{cmd}}
+
\boldsymbol{\tau}_t^{g},
\label{eq:osc_execution}
\end{equation}
where $\boldsymbol{\Lambda}_t$ is the operational-space inertia, $\mathbf J_t$ is the task Jacobian, and $\boldsymbol{\tau}_t^{g}$ is the gravity-compensation term. Equations~\eqref{eq:osc_motion_law} and \eqref{eq:osc_execution} are evaluated at the 120\,Hz physics rate, whereas the actor produces a new action every second physics step, that is, at 60\,Hz; the applied gain and the desired pose are held constant between policy steps. Controller and joint-torque rate limits are subsequently applied as described in Section~\ref{subsec:experimental_setup}.

Equations~\eqref{eq:osc_motion_law} and \eqref{eq:osc_execution} preserve the spring-damper interpretation of variable impedance while matching the implemented controller. In particular, $K_t$ is an acceleration-domain proportional gain: it modulates the closed-loop stiffness of the robot but should not be interpreted as a calibrated Cartesian spring constant in $\mathrm{N/m}$.

\subsection{Constraint-Grounded Reward}
\label{subsec:reward_design}

All force-dependent terms use a common peg--hole contact signal. At each physics substep, the pair-filtered normal and friction resultants are summed and projected onto the insertion axis $\hat{\mathbf d}$ to implement \eqref{eq:axial_force}. Filtering by the peg--hole pair excludes forces generated by the robot grasp. The resulting quantity is a simulation estimate of the axial fixture reaction rather than a wrist-wrench measurement.

The reward combines task progress, force regulation, advance-rate adaptation, action regularization, and terminal outcome:
\begin{equation}
\begin{aligned}
r_t
={}&
r_t^{\mathrm{task}}
-
w_F\ell_t^F
-
w_{\mathrm{rb}}\ell_t^{\mathrm{rb}}
\\
&
-
w_{\mathrm{proj}}\ell_t^{\mathrm{proj}}
-
w_{\mathrm{res}}\ell_t^{\mathrm{res}}
+
r_t^{\mathrm{term}},
\end{aligned}
\label{eq:reward_total}
\end{equation}
where the nonnegative weights are fixed across the complete training distribution.

The task term is based on a bounded geometric potential $\Phi_t\in[0,1]$ that combines lateral alignment, angular alignment, and insertion depth. Its temporal change rewards geometric progress, and a stage cost discourages unnecessary delay and hovering away from the goal:
\begin{equation}
r_t^{\mathrm{task}}
=
w_{\Delta\Phi}
\left(
\Phi_t-\Phi_{t-1}
\right)
-
w_{\mathrm{stage}}
\left(
1-\Phi_t
\right),
\label{eq:task_reward}
\end{equation}
where $w_{\Delta\Phi}$ and $w_{\mathrm{stage}}$ are fixed across all task specifications. The geometric potential does not depend on $F_{\max}$ or $\mathcal K$.

Let $F_{t,j}^{\mathrm{ax}}$ denote the axial reaction at physics substep $j\in\{1,\ldots,D\}$ within policy interval $t$.

A normalized squared hinge provides a smooth penalty as the contact reaction approaches the force limit:
\begin{equation}
\ell_t^F
=
\frac{1}{D}
\sum_{j=1}^{D}
\left[
\frac{
F_{t,j}^{\mathrm{ax}}
-
(1-\rho_F)F_{\max}
}{
F_{\max}
}
\right]_+^2,
\label{eq:force_margin_penalty}
\end{equation}
where $\rho_F\in(0,1)$ defines the margin below the limit at which the penalty becomes active. Normalization by $F_{\max}$ expresses the penalty relative to the force limit of the current task.

To regulate the insertion tempo, we compute the root-mean-square (RMS) force utilization over the same policy interval and define a force-limit-conditioned rate target:
\begin{align}
\overline F_t
&=
\sqrt{
\frac{1}{D}
\sum_{j=1}^{D}
\left(
F_{t,j}^{\mathrm{ax}}
\right)^2
},
\qquad
\chi_t
=
\operatorname{clip}
\left(
\frac{\overline F_t}{F_{\max}},
0,1
\right),
\nonumber\\
b(F_{\max})
&=
2
\frac{
F_{\max}-F_{\mathcal F}^{\min}
}{
F_{\mathcal F}^{\max}-F_{\mathcal F}^{\min}
}
-1,
\nonumber\\
a_t^{v,*}
&=
\operatorname{clip}
\left(
1-\lambda_u \chi_t+\lambda_b b(F_{\max}),
-1,1
\right),
\label{eq:advance_rate_target}
\end{align}
where $F_{\mathcal F}^{\min}$ and $F_{\mathcal F}^{\max}$ delimit the training range of the force limit. The utilization term lowers the target rate as the axial reaction approaches $F_{\max}$, whereas the force-limit coordinate $b(F_{\max})$ partially offsets that reduction when a larger load is allowed. The target only specifies the intended direction of rate adaptation; whether the trained actor actually exhibits this response in closed loop is examined in Section~\ref{subsec:q3_results}.

The rate-tracking term is active only in policy intervals in which the axial reaction exceeds $\epsilon_F$ and is
combined with a physics-substep penalty near the force limit:
\begin{equation}
\begin{aligned}
\ell_t^{\mathrm{rb}}
={}&
\mathbb I
\left[
\max_j F_{t,j}^{\mathrm{ax}}>\epsilon_F
\right]
\frac{1}{2}
\left(
\frac{
a_t^v-a_t^{v,*}
}{2}
\right)^2
\\
&+
\lambda_{\mathrm{bar}}
\frac{1}{D}
\sum_{j=1}^{D}
\left[
\frac{
F_{t,j}^{\mathrm{ax}}/F_{\max}-\alpha_F
}{
1-\alpha_F
}
\right]_+^2,
\end{aligned}
\label{eq:rate_barrier_penalty}
\end{equation}
where $\mathbb I[\cdot]$ is the indicator function, $\epsilon_F$ is the contact threshold, and $\alpha_F\in(0,1)$ is the utilization at which the near-limit penalty begins. Although the second term is barrier-shaped, it is a soft penalty rather than a control barrier function and does not by itself guarantee $F_{t,j}^{\mathrm{ax}}\leq F_{\max}$. The two force terms play complementary roles: the margin term $\ell_t^F$ provides broad shaping before the prescribed limit is reached, whereas the second term in $\ell_t^{\mathrm{rb}}$ penalizes short, high-utilization transients close to the limit. The force margin, rate target, barrier, and reward coefficients used in \eqref{eq:reward_total}--\eqref{eq:gain_projection_penalty} are listed in Table~\ref{tab:implementation_constants}.

The projection penalty is
\begin{equation}
\ell_t^{\mathrm{proj}}
=
\left(
\frac{
K_t^{\mathrm{req}}
-
K_t^{\mathrm{proj}}
}{
K_{\mathrm{sys}}^{\max}
-
K_{\mathrm{sys}}^{\min}
}
\right)^2.
\label{eq:gain_projection_penalty}
\end{equation}
This penalty does not reveal the gain-set endpoints to the actor; it only discourages raw requests that are repeatedly clipped under the sampled training tasks. Because the actor observes the applied gain and the contact response, it can adjust its subsequent requests through interaction feedback. No midpoint reference, phase-dependent gain schedule, or force-derived gain target is provided. The gain request is instead learned from the task reward, force response, and the consequences of projection. Finally, successful insertion receives a positive terminal impulse, whereas timeout, numerical failure, or termination by a fixed physical guard receives a negative terminal impulse. The physical guard is independent of the task-specific force limit and only prevents unreasonable simulation states; it is not counted as enforcement of $F_{\max}$.

The residual action penalty regularizes the learned pose residual,
\begin{equation}
\ell_t^{\mathrm{res}}
=
\left\|
a_t^{\mathrm{res}}
\right\|_2^2 ,
\label{eq:residual_action_penalty}
\end{equation}
which discourages large corrections to the nominal primitive without constraining their direction.

\subsection{Training and Deployment}
\label{subsec:training_deployment}

At the beginning of each training episode, the force limit, the admissible gain set, and the friction condition are sampled independently and held fixed for the episode. A single actor is trained with proximal policy optimization (PPO) across the resulting distribution of task specifications and contact conditions. Section~\ref{subsec:experimental_setup} describes the task sampling distributions. Table~\ref{tab:implementation_constants} summarizes the fixed controller, reward, and PPO constants. Table~\ref{tab:setup_parameters} reports the network architecture and training budget.

At deployment, $F_{\max}$ and $\mathcal K$ are supplied as task requirements, and no other controller parameters are changed. The actor runs without online gradient updates, friction estimation, or material-specific policy selection; its response to unobserved friction arises solely through interaction feedback.

For analysis, the implementation records the raw gain and advance-rate actions, the requested gain $K_t^{\mathrm{req}}$, the projected gain $K_t^{\mathrm{proj}}$, the applied gain $K_t$, and the resulting force, motion, and task outcome. These signals support three types of tests. Varying $F_{\max}$ while holding $\mathcal K$ fixed tests the response to the permitted contact load. Varying $\mathcal K$ at fixed $F_{\max}$ tests the response to the available controller authority. Varying friction at a fixed task specification tests the closed-loop response to the hidden contact condition.

\section{Experimental Evaluation}
\label{sec:experiments}

\begin{table*}[!t]
\centering
\caption{\textbf{Evaluation protocol and episode accounting for the four studies.}}
\label{tab:evaluation_protocol}
\footnotesize
\setlength{\tabcolsep}{3.5pt}
\renewcommand{\arraystretch}{1.12}
\resizebox{\textwidth}{!}{%
\begin{tabular}{llllr}
\toprule
Study
& Evaluation conditions
& Methods
& Primary evidence
& Episodes \\
\midrule
Study I
& $3$ force limits $\times$ $3$ gain sets $\times$ $4$ friction values
  ($\mu_s\in\{0.60,0.85,0.95,1.10\}$)
& All methods in Table~\ref{tab:compared_methods}
& Constraint-compliant success and task efficiency
& 28{,}800 \\
Study II
& $7$ contact and observation shifts $\times$ $3$ force limits
  $\times$ $3$ gain sets
& All methods in Table~\ref{tab:compared_methods}
& Robustness and operating boundaries
& 50{,}400 \\
Study III
& $11$ force limits from $6.5$ to $9.0$\,N,
  with friction and gain set fixed
& CG-RL, No $F_{\max}$ input, and $K$-conditioned
& Raw advance-action slope
& 5{,}280 \\
Study IV
& $6$ force limits sampled from
  $[5.5,6.25]\cup[9.25,10.0]$\,N
& CG-RL
& Extrapolation and failure boundary
& 960 \\
\bottomrule
\end{tabular}%
}
\end{table*}

The evaluation comprises four studies, summarized in Table~\ref{tab:evaluation_protocol}. Study I compares CG-RL with controlled ablations and a fixed-gain baseline. Study II evaluates the robustness of CG-RL to unobserved contact variation and sensing perturbations. Study III examines whether the actor responds continuously to the prescribed force limit. Study IV evaluates generalization beyond the force-limit range used during training. Across these studies, policy behavior, gain enforcement, and contact response are analyzed separately.
\subsection{Experimental Setup}
\label{subsec:experimental_setup}

\subsubsection{Task and Simulator}
\begin{figure*}[t]
    \centering
    \includegraphics[width=\textwidth]
    {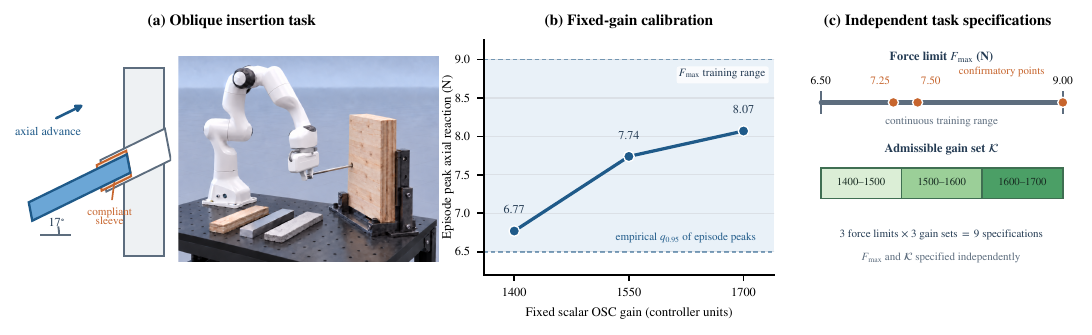}
    \caption{\textbf{Task, calibration, and specifications.}
(a) The 17$^\circ$ oblique insertion geometry and an illustrative
simulation view; the material samples are shown for context.
(b) Empirical 95th percentile of episode peak axial reaction from
21 successful calibration trials at each fixed gain; the dashed bounds
mark the force-limit training range.
(c) Confirmatory force limits and three admissible gain sets, varied
independently. The task schematic is not drawn to scale.}
    \label{fig:task_calibration}
\end{figure*}

The experiment is motivated by oblique fastening and constrained assembly, where geometric misalignment and inclined contact can generate substantial lateral and axial reactions. As shown in Fig.~\ref{fig:task_calibration}(a), the task is implemented in Isaac Lab using a Franka Research~3 robot and Isaac Sim contact dynamics. The robot inserts a peg into a bore tilted by approximately $17^\circ$. A short compliant sleeve inside the bore produces a repeatable contact load; it serves as a simulation contact proxy rather than as a constitutive model of a particular material.

A trial is geometrically successful when the peg reaches an insertion depth of 30\,mm and remains within the terminal pose tolerance for five consecutive policy steps. Each episode represents one insertion attempt and terminates upon success, failure, or timeout. Table~\ref{tab:setup_parameters} summarizes the task geometry, simulation rates, episode horizon, and training perturbations.

\begin{table}[t]
\centering
\caption{Key simulation, training, and evaluation settings.}
\label{tab:setup_parameters}
\footnotesize
\setlength{\tabcolsep}{3pt}
\renewcommand{\arraystretch}{1.10}

\begin{tabular}{@{}p{0.36\columnwidth}p{0.58\columnwidth}@{}}
\toprule
Setting & Value \\
\midrule

Robot
& Franka Research~3 \\

Simulation / policy rate
& 120 / 60\,Hz ($D=2$ physics substeps per policy step) \\

Simulator
& Isaac Sim 5.0.0, Isaac Lab 2.2.1 \\

PhysX solver
& TGS; 8 position iterations and 1 velocity iteration for the robot,
  192 position iterations and 1 velocity iteration for the peg and bore \\

Episode horizon
& 25\,s \\

Target insertion depth
& 30\,mm \\

Force limit sampling
& $\mathcal{U}[6.5,7.0]$ ($p=0.4$),
  $\mathcal{U}[7.0,8.0]$ ($p=0.3$),
  $\mathcal{U}[8.0,9.0]$ ($p=0.3$)\,N \\

Gain sets
& $[1400,1500]$, $[1500,1600]$,
  $[1600,1700]$ \\

Static friction sampling
& $\mathcal{U}[0.60,0.85]$ ($p=0.4$),
  $\mathcal{U}[0.85,1.10]$ ($p=0.6$) \\

Parallel environments
& 512 \\

Training algorithm
& Proximal policy optimization (PPO) \\

Training iterations
& 600 \\

Actor and critic network
& Multilayer perceptron (MLP),
  $[512,128,64]$ \\

Hidden activation
& Exponential linear unit (ELU) \\

Independent training seeds
& 5 \\

\bottomrule
\end{tabular}
\end{table}

\begin{table*}[t]
\centering
\caption{\textbf{Fixed controller, reward, and PPO constants.} Reward coefficients are reported as nonnegative magnitudes; their signs follow \eqref{eq:reward_total}--\eqref{eq:gain_projection_penalty}. All learned methods use the same values unless stated otherwise.}
\label{tab:implementation_constants}
\footnotesize
\setlength{\tabcolsep}{4pt}
\renewcommand{\arraystretch}{1.12}
\begin{tabular}{llll}
\toprule
Parameter & Value & Parameter & Value \\
\midrule

Advance rate bounds $(s_{\min},s_{\max})$
& $(0.5,1.5)$
& System gain range $\mathcal{K}_{\mathrm{sys}}$
& $[1400,1700]$ \\

Maximum gain rate $\dot K_{\max}^{\mathrm{cmd}}$
& $2800$ controller units/s
& Fixed physical guard
& $75$\,N \\

Force margin $\rho_F$
& $0.10$
& Contact threshold $\epsilon_F$
& $0.5$\,N \\

Rate target coefficient $\lambda_u$
& $2.0$
& Force limit coefficient $\lambda_b$
& $0.75$ \\

Barrier onset $\alpha_F$
& $0.80$
& Barrier scale $\lambda_{\mathrm{bar}}$
& $0.20$ \\

Force limit range $(F_{\mathcal F}^{\min},F_{\mathcal F}^{\max})$
& $(6.5,9.0)$\,N
& Progress and stage weights $(w_{\Delta\Phi},w_{\mathrm{stage}})$
& $(5,1)$ \\

Force and rate weights $(w_F,w_{\mathrm{rb}})$
& $(100,5)$
& Projection and residual weights $(w_{\mathrm{proj}},w_{\mathrm{res}})$
& $(25,1)$ \\

Terminal impulses (success, failure)
& $(+5,-1)$
& Discount factor $\gamma$
& $0.999$ \\

Generalized advantage estimation (GAE) parameter
$\lambda_{\mathrm{GAE}}$
& $0.95$
& Learning rate
& $3\times10^{-4}$, fixed \\

PPO clip parameter
& $0.20$
& Entropy coefficient
& $0$ \\

Critic coefficient
& $2.0$
& Gradient norm limit
& $1.0$ \\

Rollout length
& $128$ steps
& Minibatch size
& $512$ \\

PPO epochs per iteration
& $4$
& Action clipping
& $[-1,1]$ \\

\bottomrule
\end{tabular}
\end{table*}

\subsubsection{Operating Range Calibration}
The force limits and gain ranges used in training and evaluation were determined by a fixed-controller calibration performed before any learned method was compared. The operating range was therefore defined from measured contact responses rather than from policy results. Calibration used the three fixed scalar gains shown in Fig.~\ref{fig:task_calibration}(b) and crossed the prescribed pose and friction conditions, giving 21 trials at each gain. All perturbations remained within the training range. The requested and applied gains were held constant from the first control step, while the nominal motion primitive and all other controller settings remained unchanged.

For each trial, we recorded the maximum axial reaction over the complete episode and summarized the trial-level maxima by their empirical 0.95 quantile. All calibration trials completed successfully. As shown in Fig.~\ref{fig:task_calibration}(b), this upper-tail statistic increased by 1.30\,N from the lowest to the highest gain, whereas the median changed little. These measurements are used only to select the operating range for the subsequent experiments and are not interpreted as precise probabilistic force boundaries.

The gain interval $[1400,1700]$ was selected as a stable operating window for the present task. All three calibration gains completed the prescribed conditions while producing a measurable change in controller authority and contact loading. The interval is not unique, and its endpoints are not mechanical limits of the robot.

The same measurements were used to set the force-limit training range to $F_{\max}\in[6.5,9.0]$\,N. This interval spans restrictive and permissive conditions relative to the calibrated upper-tail reactions. During training, the force limit and friction are sampled independently from the nonuniform mixtures in Table~\ref{tab:setup_parameters}. The three confirmatory limits shown in Fig.~\ref{fig:task_calibration}(c) probe restrictive, transitional, and permissive behavior. A separate development pilot verified this coverage; its episodes are not included in the confirmatory results.

For the factorial study, the controller interval is divided into the three adjacent equal-width gain sets shown in Fig.~\ref{fig:task_calibration}(c). This partition provides ordered, balanced controller conditions while preserving continuous coverage of the calibrated interval. The force limit and gain set are specified independently.

\subsubsection{Training and Compared Methods}
All learned methods use the same task, action space, controller, PPO implementation, training distribution, network architecture, and optimization budget. Each method is trained from random initialization under five independent seeds using the settings in Tables~\ref{tab:setup_parameters} and~\ref{tab:implementation_constants}. No intermediate checkpoint selection is performed; all evaluations use the final checkpoint after 600 iterations.

\begin{table}[t]
\centering
\caption{Methods included in the main comparison.}
\label{tab:compared_methods}
\footnotesize
\setlength{\tabcolsep}{2.5pt}
\renewcommand{\arraystretch}{1.08}
\resizebox{\columnwidth}{!}{%
\begin{tabular}{llccc}
\toprule
Role
& Method
& Task input to actor
& Obs. dim.
& Gain execution \\
\midrule

Proposed
& CG-RL
& $F_{\max}$
& 35
& Set projection \\

Controlled ablation
& No $F_{\max}$ input
& None
& 34
& Set projection \\

Learned comparator
& Margin barrier
& $F_{\max},K_{\min},K_{\max}$
& 37
& Set projection \\

Descriptive comparator
& $K$-conditioned
& $K_{\min},K_{\max}$
& 36
& Set projection \\
Fixed baseline
& Fixed midpoint controller
& None
& --
& Fixed midpoint controller \\

\bottomrule
\end{tabular}%
}
\end{table}

CG-RL supplies $F_{\max}$ to the actor, while the gain set is provided separately to the execution layer. The controlled ablation removes $F_{\max}$ from the actor observation while preserving the policy architecture, reward, controller, training distribution, and gain set projection.

Margin barrier replaces the proposed force regulation objective with a physics step margin penalty. It also observes the complete task descriptor and is therefore treated as an alternative learned comparator rather than a controlled ablation. The fixed midpoint controller is a deterministic baseline that applies the midpoint of the supplied gain set throughout the episode.

\subsubsection{Evaluation Protocol}
All evaluations use frozen checkpoints and fresh episode banks that were not used for training, checkpoint selection, or method selection. Within each condition, all methods receive the same reset geometry, task specification, and contact parameters at each episode index. The initial conditions therefore remain paired across methods, even though the resulting trajectories differ. For each learned method, every seed and condition cell contains 32 episodes. Each condition cell for a fixed-baseline evaluation block also contains 32 episodes.

The primary metric is constraint-compliant success (CCS). A trial counts as a constraint-compliant success only if the insertion is geometrically completed, the axial reaction satisfies $F_t^{\mathrm{ax}}\leq F_{\max}$ at every physics step, and the applied gain remains inside the supplied set throughout the episode. For the projected methods, gain-set membership is enforced by the execution layer, so CCS evaluates the complete actor--controller system rather than treating gain membership as a learned property.

Two further definitions are used throughout. The prespecified stress subset contains the conditions with $F_{\max}\leq7.50$\,N and $\mu_s\geq0.85$. Failure-aware completion time assigns the full 25\,s horizon to every episode that is not a constraint-compliant success. This prevents methods with many early failures from appearing artificially fast. An episode that completes the insertion but violates the force limit is also censored at the horizon. The metric is therefore horizon-censored rather than a conventional completion time and remains comparable because every method uses the same horizon.

For learned methods, the five training seeds are the statistical units; individual episodes are not treated as independent samples. We report the mean, sample standard deviation, and two-sided 95\% Student's $t$ confidence interval. Paired $t$ tests are primary, with exact two-sided sign-flip tests used as small-sample checks; Holm correction is applied to the paired-$t$ $p$ values within each family. Learned methods are paired by seed. For the fixed baseline, each CG-RL seed is paired with a matched evaluation-bank block; these five blocks are not independent training replications.

The locked Study I family contains four stress-subset CCS contrasts: CG-RL versus No $F_{\max}$ input, Margin barrier, $K$-conditioned, and the fixed midpoint controller. The $K$-conditioned comparator remains in this family but is interpreted descriptively because it changes both descriptor visibility and observation dimension. The initial Study III sweep used $K$-conditioned; the No $F_{\max}$ sweep reported here was added with frozen checkpoints and matched banks as a controlled conditioning analysis.
\subsection{Task Performance and Ablations}
\label{subsec:q1_results}
Study I compares the five methods in Table~\ref{tab:compared_methods} under the training distribution and then isolates the contribution of online gain selection through two paired interventions.
\subsubsection{Task Performance Comparison}
\begin{figure*}[!t]
\centering
\includegraphics[
    width=0.84\textwidth,
    height=0.43\textheight,
    keepaspectratio
]{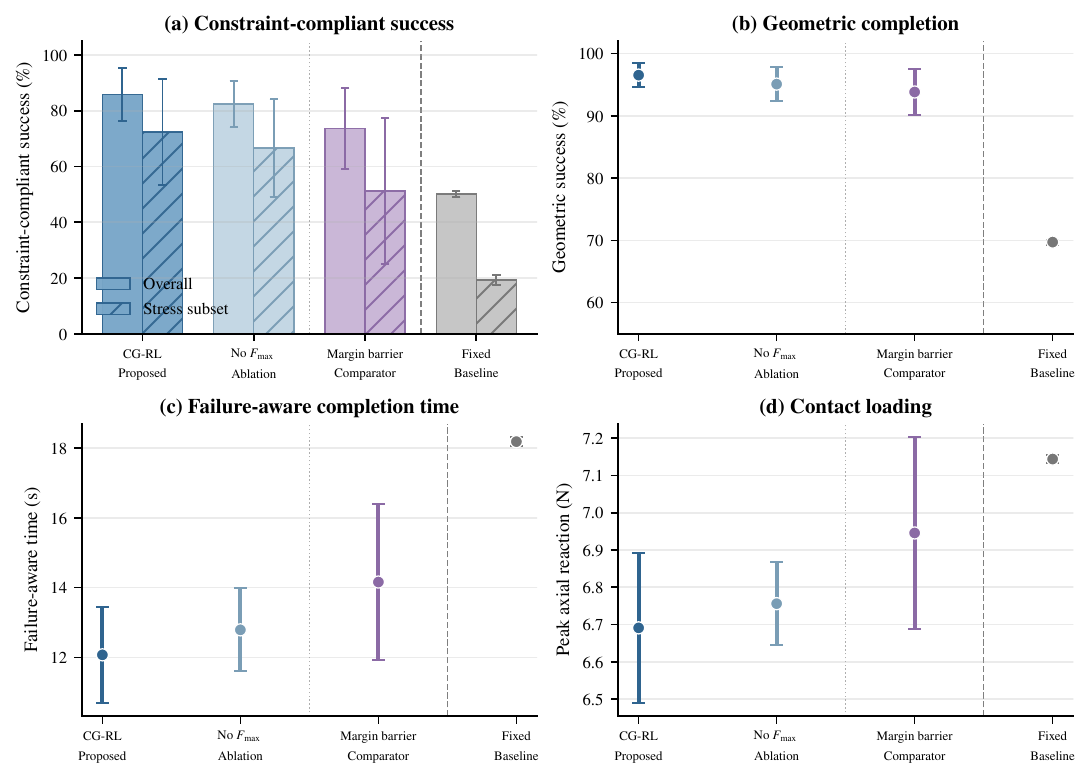}
\caption{\textbf{Task performance under the training distribution.}
CG-RL is the proposed method. The light blue entry is the controlled force limit ablation, the purple entry is the learned objective comparator, and the
gray entry is the fixed midpoint controller. Error bars show 95\% confidence intervals across five independent training seeds for the learned methods and five paired evaluation-bank blocks for the fixed baseline.}
\label{fig:q1_results}
\end{figure*}

Table~\ref{tab:q1_results} and Fig.~\ref{fig:q1_results} report the Study I results. For each learned method, the metrics are computed separately for the five training seeds and then summarized across seeds. The fixed midpoint controller is summarized over five paired evaluation-bank blocks.

\begin{table*}[t]
\centering
\caption{\textbf{Task performance under the training distribution.}}
\label{tab:q1_results}
\setlength{\tabcolsep}{5pt}
\renewcommand{\arraystretch}{1.12}
\resizebox{\textwidth}{!}{%
\begin{tabular}{llccccc}
\toprule
Role
& Method
& Overall CCS [\%]
& Stress-subset CCS [\%]
& Geometric success [\%]
& Peak axial reaction [N]
& Completion time [s] \\
\midrule
Proposed
& CG-RL
& 85.78 $\pm$ 7.73
& 72.33 $\pm$ 15.28
& 96.53 $\pm$ 1.51
& 6.691 $\pm$ 0.162
& 12.08 $\pm$ 1.11 \\
\midrule
Controlled ablation
& No $F_{\max}$ input
& 82.43 $\pm$ 6.71
& 66.63 $\pm$ 14.03
& 95.09 $\pm$ 2.23
& 6.756 $\pm$ 0.089
& 12.79 $\pm$ 0.96 \\
Learned comparator
& Margin barrier
& 73.58 $\pm$ 11.73
& 51.25 $\pm$ 21.03
& 93.82 $\pm$ 3.00
& 6.946 $\pm$ 0.208
& 14.16 $\pm$ 1.81 \\
Descriptive comparator
& $K$-conditioned
& 89.34 $\pm$ 2.45
& 80.52 $\pm$ 4.02
& 95.28 $\pm$ 3.59
& 6.559 $\pm$ 0.179
& 11.71 $\pm$ 0.51 \\
\midrule
Fixed baseline
& Fixed midpoint controller
& 50.10 $\pm$ 0.96
& 19.34 $\pm$ 1.49
& 69.72 $\pm$ 0.32
& 7.144 $\pm$ 0.008
& 18.18 $\pm$ 0.11 \\
\bottomrule
\end{tabular}%
}
\end{table*}

Relative to the fixed midpoint controller, CG-RL raises stress-subset CCS from 19.3\% to 72.3\%, a 53-percentage-point improvement (Holm-adjusted paired $t$ test, $p=0.0075$). The exact sign-flip test gives its minimum possible two-sided value with five pairs but does not reach 0.05. CG-RL also achieves higher geometric completion and shorter failure-aware completion time.

Removing $F_{\max}$ produces modest mean declines in overall and stress-subset CCS. Margin barrier declines further, whereas $K$-conditioned has the highest aggregate CCS. None of the learned comparators differs significantly from CG-RL after correction. Aggregate metrics alone therefore cannot establish whether the actor responds to the prescribed force limit.

\begin{figure*}[t]
    \centering
    \includegraphics[width=\textwidth]
    {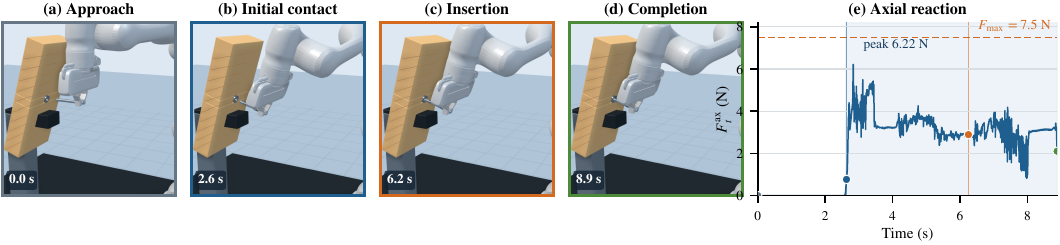}
    \caption{\textbf{Representative CG-RL rollout.}
    A successful stress-subset episode at $F_{\max}=7.5$\,N,
    $\mathcal K=[1500,1600]$, and $\mu_s=0.85$.
    }
    \label{fig:representative_rollout}
\end{figure*}

Study I therefore yields two findings. CG-RL substantially outperforms the fixed midpoint controller, whereas differences among the learned methods remain inconclusive. Whether the actor actually changes its behavior with $F_{\max}$ is examined directly in Section~\ref{subsec:q3_results}. Figure~\ref{fig:representative_rollout} shows one successful stress-subset episode and its axial-force response.

\subsubsection{Online Gain Selection and Gain Set Enforcement}
\label{subsubsec:gain_intervention}
\begin{figure*}[!t]
    \centering
    \includegraphics[width=\textwidth]
    {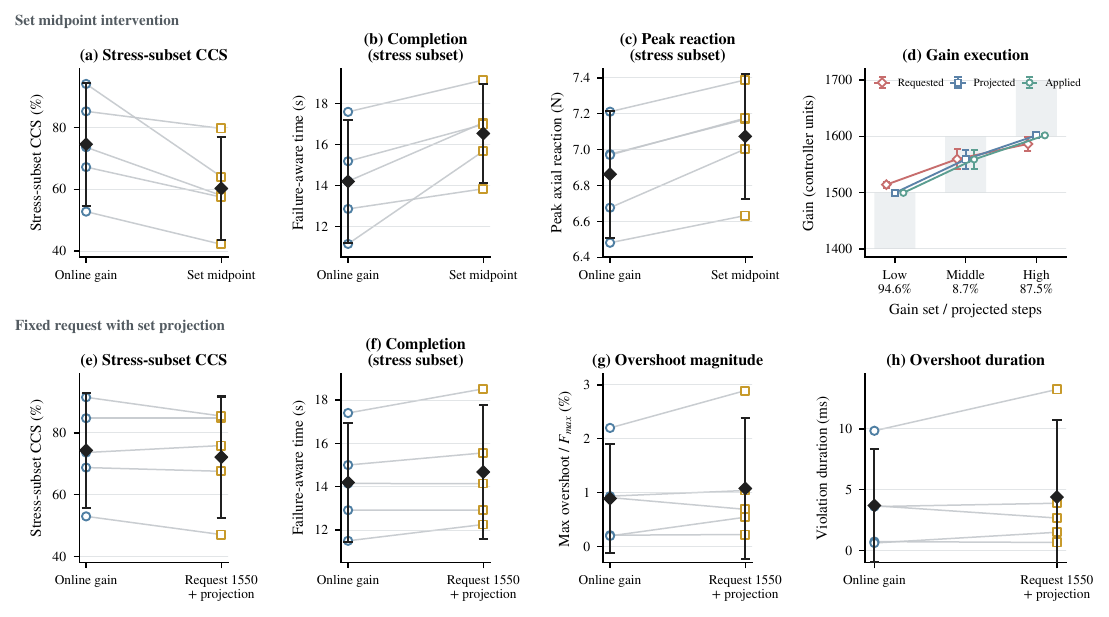}
    \caption{\textbf{Online gain selection and gain set enforcement.}
Panels (a)--(c) evaluate the set midpoint intervention, panel (d)
reports requested, projected, and applied gains, and panels (e)--(h)
evaluate a fixed gain request of 1550 under set projection.
Percentages in panel (d) indicate the fraction of contact steps modified
by projection. Each gray line connects the paired results for one training seed; black diamonds and error
bars show the mean and two-sided 95\% confidence interval across five seeds.}
    \label{fig:gain_contribution_execution}
\end{figure*}

We next examine the gain channel through two paired interventions. In the standard mode, the actor updates its gain request throughout the episode while also commanding the residual pose and advance rate. In both interventions, these motion commands continue to be computed from the current observation, and only the gain request is replaced.
The first intervention fixes the gain at the midpoint of the supplied set for the complete episode,
\[
K_t = K_{\mathrm{mid}}, \qquad t=0,\ldots,T-1.
\]
This intervention removes online gain adjustment as well as any variation of the gain within a set. The standard and intervened executions use the same policy checkpoints and paired episode banks. The second intervention provides a more tightly controlled comparison. It replaces the actor request with the constant value 1550 but retains the same set projection and gain rate limit used by CG-RL:
\[
K_t^{\mathrm{req}} = 1550,
\qquad
K_t^{\mathrm{proj}} = \Pi_{\mathcal K}(1550).
\]
Consequently, the request is projected to the nearest boundary for the low and high sets but remains unchanged for the middle set. This intervention removes feedback-based gain selection while preserving the response of the execution layer to the supplied gain set. Fig.~\ref{fig:gain_contribution_execution} reports both interventions together with the gain execution path under standard CG-RL.

Each intervention uses a separate paired episode bank, with standard and intervened executions matched episode by episode. The reference values for standard CG-RL therefore differ slightly across banks and from Table~\ref{tab:q1_results}. These differences are small relative to the seed-to-seed variation and do not affect the paired interpretation.

Fixing the gain at the midpoint reduces stress-subset CCS by 14.31 percentage points (95\% confidence interval $[2.52,26.09]$ percentage points). Failure-aware completion time and peak axial reaction also increase, as shown in Fig.~\ref{fig:gain_contribution_execution}. Because the policy weights, episode conditions, residual pose commands, and advance-rate commands remain unchanged, the decline can be attributed to replacing the online gain command with $K_{\mathrm{mid}}$.

The constant-request intervention produces smaller changes. Fixing the request at 1550 before projection slightly reduces stress-subset CCS and increases completion time, normalized overshoot, and violation duration. Here, maximum normalized overshoot is the episode peak of $v_t^{F}$ in \eqref{eq:force_violation}, divided by $F_{\max}$. None of the paired differences is statistically significant across the five training seeds. A constant request combined with set projection therefore recovers most, but not all, of the performance of online gain selection.

Fig.~\ref{fig:gain_contribution_execution}(d) shows how the execution layer treats the actor requests. The mean request shifts only modestly with the supplied set and stays within a narrow band around 1550. For the low and high sets this request lies outside the set, so projection is active during most contact-phase steps and the applied gain sits at the
nearer set boundary. For the middle set the request lies inside the set and projection rarely intervenes. At the set boundaries, the supplied set rather than the actor request determines the applied gain.

The two interventions isolate the roles of online gain selection and set projection. The midpoint intervention shows that a constant gain at the center of each set does not reproduce the performance of online gain selection. The constant-request intervention shows that projecting a common request accounts for much of the response to the supplied gain set. The actor adjusts its gain request using contact feedback, while the execution layer keeps the applied gain within the supplied set. However, these experiments do not show that online gain selection outperforms every fixed request combined with projection.
\subsection{Robustness and Operating Boundaries}
\label{subsec:q2_results}
Study II examines two forms of robustness. We first trace the response of CG-RL across a continuous range of friction coefficients and then compare all methods under separate changes in friction, contact properties, fixture pose, and observations.
\subsubsection{Response to Friction Variation}
\label{subsubsec:q2_friction}
Because the friction coefficient is not part of the actor observation, this experiment measures how the frozen CG-RL policies respond through interaction feedback when the contact resistance changes. Results at each friction value are averaged over the full factorial set of confirmatory force limits and gain sets. Static friction is varied while the dynamic-to-static ratio and all other conditions remain fixed.
\begin{figure*}[t]
    \centering
    \includegraphics[width=\textwidth]
    {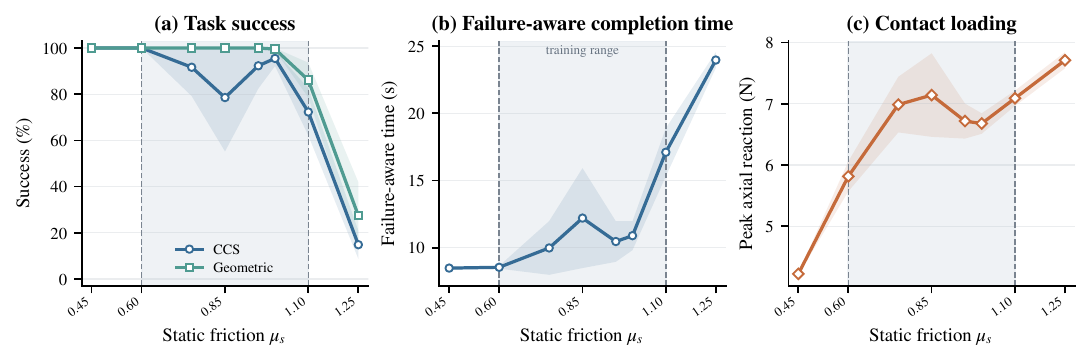}
    \caption{\textbf{Response of CG-RL to static friction.}
    The shaded interval denotes the friction range used during training. The panels report constraint-compliant and geometric success, failure-aware completion time, and mean episode peak axial reaction. Each friction value summarizes three force limits, three gain sets, and 32 paired episodes in each factorial cell. Curves and bands show the mean and two-sided 95\% confidence interval across five independent training seeds.}
    \label{fig:q2_results}
\end{figure*}
As shown in Fig.~\ref{fig:q2_results}, CCS remains above 72\% throughout the training range, while geometric success is nearly saturated over most of this interval. Completion time generally increases as contact resistance becomes more severe, although CCS is not strictly monotonic within the training range.

Performance remains high at the lower extrapolation point. At the upper point, $\mu_s=1.25$, CCS falls to 14.9\% and completion time approaches the episode horizon. This condition marks a high-resistance operating boundary characterized by slower and less reliable insertion under increased contact loading.

\subsubsection{Contact and Observation Shifts}
\label{subsubsec:q2_shifts}
We next compare all methods under the seven prespecified shifts in Table~\ref{tab:q2_results}. The friction columns reuse the extrapolation endpoints in Fig.~\ref{fig:q2_results}; the remaining columns vary contact compliance, fixture pose, or sensor noise. The combined condition applies the most demanding shifts simultaneously.
\begin{table*}[t]
\centering
\caption{\textbf{Constraint-compliant success under contact and observation shifts, reported as mean $\pm$ sample standard deviation (\%).}}
\label{tab:q2_results}
\footnotesize
\setlength{\tabcolsep}{4pt}
\renewcommand{\arraystretch}{1.12}
\resizebox{\textwidth}{!}{%
\begin{tabular}{llccccccc}
\toprule
Role
& Method
& Low friction
& High friction
& Soft contact
& Stiff contact
& \shortstack{Fixture pose\\variation}
& Sensor noise
& Combined \\
&
& $\mu_s=0.45$
& $\mu_s=1.25$
& $\times 0.75$
& $\times 1.25$
& \shortstack{$2\times$ nominal\\reset range}
& Bounded
& All shifts \\
\midrule
Proposed
& CG-RL
& 100.00 $\pm$ 0.00
& 14.86 $\pm$ 4.91
& 94.79 $\pm$ 6.07
& 32.50 $\pm$ 7.50
& 84.58 $\pm$ 7.71
& 94.44 $\pm$ 6.76
& 0.00 $\pm$ 0.00 \\
Controlled ablation
& No $F_{\max}$ input
& 100.00 $\pm$ 0.00
& 15.83 $\pm$ 1.46
& 92.71 $\pm$ 8.02
& 29.31 $\pm$ 5.38
& 82.57 $\pm$ 4.49
& 91.53 $\pm$ 5.67
& 0.00 $\pm$ 0.00 \\
Learned comparator
& Margin barrier
& 100.00 $\pm$ 0.00
& 14.79 $\pm$ 3.62
& 91.25 $\pm$ 7.57
& 26.04 $\pm$ 4.03
& 71.67 $\pm$ 12.54
& 78.82 $\pm$ 15.86
& 0.00 $\pm$ 0.00 \\
Descriptive comparator
& $K$-conditioned
& 100.00 $\pm$ 0.00
& 14.17 $\pm$ 6.22
& 93.68 $\pm$ 10.03
& 30.00 $\pm$ 14.93
& 85.00 $\pm$ 4.41
& 94.65 $\pm$ 3.43
& 0.00 $\pm$ 0.00 \\
\midrule
Fixed baseline
& Fixed midpoint controller
& 100.00 $\pm$ 0.00
& 0.00 $\pm$ 0.00
& 34.58 $\pm$ 3.26
& 0.69 $\pm$ 0.43
& 32.57 $\pm$ 2.06
& 34.58 $\pm$ 1.46
& 0.00 $\pm$ 0.00 \\
\bottomrule
\end{tabular}%
}
\end{table*}
CG-RL retains more than 84\% CCS under soft contact, increased fixture pose variation, and bounded sensor noise. Performance drops sharply under stiff contact and high friction, identifying these conditions as the dominant stressors. The learned comparators exhibit the same general pattern, with relatively small differences under several individual shifts. We therefore use these tests primarily to identify the operating range of the learned controllers rather than to rank them.

The fixed midpoint controller is substantially less tolerant of the individual shifts: its CCS remains below 35\% under soft contact, fixture pose variation, and sensor noise, and it nearly fails under stiff contact. The learned closed-loop controllers therefore retain useful performance over a broader range of individual changes than the fixed controller. When all shifts are applied simultaneously, however, every method records zero CCS. This common failure marks a boundary of the current task, controller, and training distribution rather than a limitation unique to CG-RL.

Tables~\ref{tab:q1_results} and~\ref{tab:q2_results} also report the $K$-conditioned comparator, whose actor observes $K_{\min}$ and $K_{\max}$ but not $F_{\max}$. Because it changes both descriptor visibility and observation dimension, it is reported descriptively rather than as a controlled ablation. It attains 89.3\% overall CCS in Study I, but its raw advance request remains essentially invariant over the Study III force-limit sweep. This result further shows that aggregate task performance and an explicit policy response to the prescribed force limit are distinct properties.
\subsection{Continuous Response to the Force Limit}
\label{subsec:q3_results}
\begin{figure*}[t]
    \centering
    \includegraphics[width=0.8\textwidth]
    {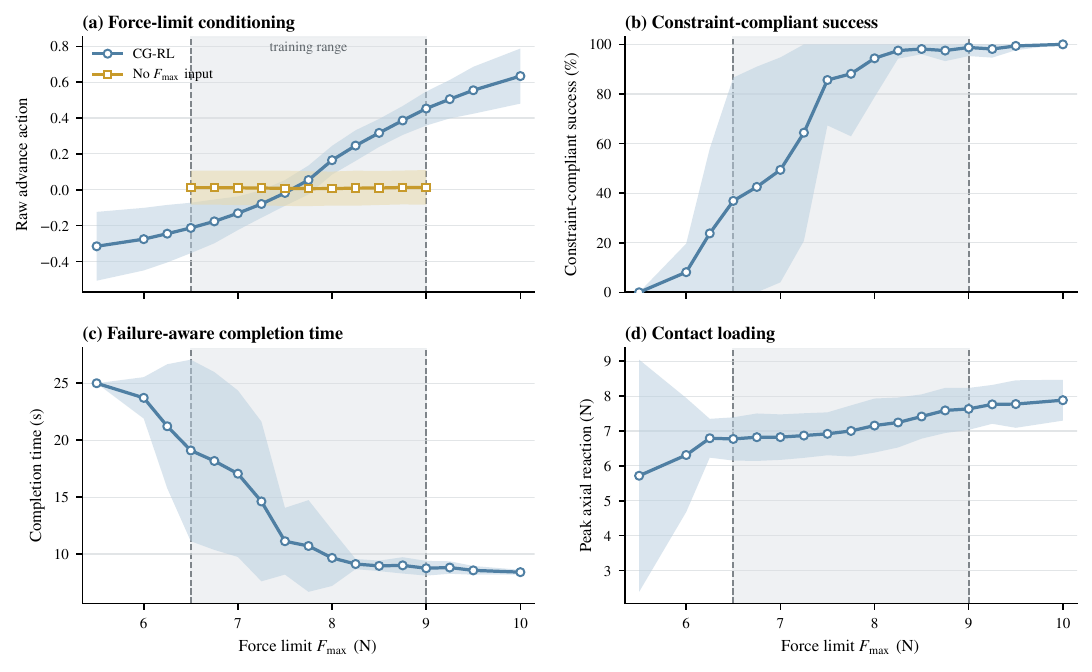}
    \caption{\textbf{Continuous response to the force limit and extrapolation beyond the training range.}
    The shaded interval denotes the force-limit range used during training.
    Panel (a) compares the contact-phase raw advance action of CG-RL with that of the No $F_{\max}$ input ablation.
    Panels (b)--(d) report the constraint-compliant success, failure-aware completion time, and peak axial reaction of CG-RL.
    Curves and bands show the mean and two-sided 95\% confidence interval across five independent training seeds.}
    \label{fig:q3_q4_results}
\end{figure*}

Study III examines whether the actor changes its behavior in response to the prescribed force limit. Aggregate success alone cannot establish this property, because a fixed policy may achieve a higher measured CCS simply because the evaluation limit is relaxed. We therefore evaluate CG-RL, the No $F_{\max}$ input ablation, and the $K$-conditioned comparator at eleven force limits spanning the training range. All policies are frozen, with $\mu_s=0.85$ and $\mathcal K=[1500,1600]$. Fig.~\ref{fig:q3_q4_results} reports the sweep; Table~\ref{tab:evaluation_protocol} provides the episode accounting.

The raw advance action is recorded before controller projection. Its dependence on $F_{\max}$ is summarized by an ordinary least-squares slope, fitted over the eleven force limits within the training range and computed separately for each training seed. Table~\ref{tab:q3_slopes} lists the resulting slopes.
\begin{table}[t]
    \centering
   \caption{\textbf{Raw advance action sensitivity to the force limit within the training range.}}
    \label{tab:q3_slopes}
    \footnotesize
    \setlength{\tabcolsep}{4pt}
    \begin{tabular}{lcc}
        \toprule
        Method
        & Mean slope [action/N]
        & Seed range [action/N] \\
        \midrule
        CG-RL
        & 0.282
        & $[0.203,\,0.368]$ \\
        No $F_{\max}$ input
        & $0.000143$
        & $[-0.000422,\,0.000597]$ \\
        \bottomrule
    \end{tabular}
\end{table}
As shown in Fig.~\ref{fig:q3_q4_results}(a), the raw advance action of CG-RL increases consistently with $F_{\max}$, whereas the No $F_{\max}$ input actor remains nearly invariant. The mean CG-RL slope is $0.282$ action/N, three orders of magnitude larger than that of the ablation. These slopes are least-squares fits over the full training-range sweep rather than differences between its endpoints.

The paired slope difference is $0.282$ action/N (95\% confidence interval $[0.195,0.369]$; paired-$t$ $p=0.000833$), with the same direction across all five seeds. The exact sign-flip test reaches its minimum two-sided value for five pairs but not 0.05. The effect size and consistent seed-level trend provide direct evidence that access to $F_{\max}$ changes the learned advance request.

The practical effect is also visible at the tightest training limit. CG-RL achieves 36.9\% CCS, compared with 3.1\% for the No $F_{\max}$ input ablation, while both methods still complete the insertion geometrically. The ablation can therefore complete the task, but it does not adjust its behavior sufficiently to satisfy the tighter force limit.

As $F_{\max}$ increases, failure-aware completion time decreases while peak axial reaction rises. Much of the time reduction follows from the higher CCS; completion time among geometrically successful trials and the mean applied gain change only marginally. The clearest learned response is therefore the change in raw advance action and the resulting use of the permitted contact force, rather than large changes in mean applied gain or successful-trial completion time.

The No $F_{\max}$ input policy remains competitive in aggregate task performance, but it does not change its raw advance action with the supplied force limit. Its measured CCS can still increase when $F_{\max}$ is relaxed, because the same contact trajectory is then evaluated against a less restrictive threshold. This comparison answers RQ2 by separating task success from an explicit policy response to the force limit.

\subsection{Extrapolation Beyond the Training Range}
\label{subsec:q4_results}
Study IV examines the response of the frozen CG-RL policies outside the force-limit range used during training. The unshaded regions of Fig.~\ref{fig:q3_q4_results} cover force limits below 6.5\,N and above 9.0\,N. The normalized force-limit observation is not clipped, no retraining or controller retuning is performed, and all other evaluation conditions remain identical to those in Study III.

Above the training range, CG-RL maintains more than 98\% CCS with complete geometric success. Its raw advance action continues to increase while failure-aware completion time decreases. The policy response therefore extends smoothly across the tested upper range without retraining or controller retuning.

The lower side reveals a different boundary. CCS reaches zero at 5.5\,N even though most episodes still complete geometrically. The raw advance action continues in the expected conservative direction. The actor therefore still responds to the prescribed limit, but the insertion cannot be completed reliably while satisfying the most restrictive force requirements.

The observed extrapolation is thus asymmetric. CG-RL remains effective above the training range, whereas force limits below approximately 6.25\,N expose a repeatable feasibility boundary for the present task and controller. These results provide simulation evidence of extrapolation above the training range over the tested force limits, but they do not establish compliance with arbitrary force limits outside the training distribution.

\section{Discussion}
\label{sec:discussion}
The experiments show that CG-RL changes its insertion behavior with the prescribed force limit, while the controller enforces the independently specified gain set. CG-RL also adapts to unobserved friction through interaction feedback. Four results clarify the origin of this behavior and identify the range over which it remains effective.

First, the gain channel is factorized as intended. Set projection accounts for most of the response to the supplied gain set, whereas online gain selection becomes valuable when contact resistance is high. Fixing the gain at the set midpoint has little effect at low friction but causes a consistent decline under high contact resistance. The actor does not observe the gain-set endpoints. However, the applied gain is part of its observation and provides indirect information about the active set. This feedback explains why the mean gain request shifts with the supplied set.

Second, aggregate task performance and an explicit response to the force limit are distinct properties. As the limit is relaxed across the Study III sweep, CG-RL uses more of the permitted load. The No $F_{\max}$ input ablation produces a nearly unchanged force trace, so its CCS improves only because the threshold is relaxed. The $K$-conditioned comparator reaches the highest overall CCS by being uniformly conservative rather than by responding to the specification. Recording the policy request, controller command, and physical response separately makes these differences visible.

Third, the operating boundaries found in Studies II and IV are strongly influenced by the mechanics of the present task and contact model. Below the training range, CG-RL continues to reduce its contact force, but the most restrictive limit cannot be satisfied reliably. At high friction, the policy slows down rather than pushing harder; this strategy remains effective within the training range but breaks down at the highest tested resistance.

Fourth, the statistical comparison is clear for the fixed controller but inconclusive among the learned methods. CG-RL substantially improves stress-subset CCS over the fixed midpoint controller, whereas the smaller differences among learned methods are not statistically significant. Direct evidence of force-limit conditioning comes from the Study III sweep, where every seed shows the same trend.

Beyond the simulated task, the specification interface of CG-RL matches how contact requirements arise in practice. In timber fastening, the allowable axial load of a joint follows from its design~\cite{Bejtka2002InclinedScrews,Wang2025CyclicGlulam}, and the admissible gain set follows from controller commissioning. Both are known before execution, whereas friction is not. A task-level planner can therefore issue both requirements per component without retraining the insertion policy.

The present study has two limitations. First, CG-RL assumes that a suitable force limit $F_{\max}$ is specified before execution. This setup allows us to isolate how the policy responds to the prescribed requirement, but the limit must still be calibrated for the target material. Second, the evaluation uses rigid peg and fixture geometry with a repeatable compliant sleeve. This setup provides controlled contact conditions for comparing methods, but it does not represent deformation of the peg or workpiece.

\section{Conclusion}
\label{sec:conclusion}
This paper presented CG-RL, a variable impedance framework for robotic insertion with a prescribed axial force limit and an independently specified controller gain set. The actor observes the force limit and contact feedback and produces a residual motion, an advance rate, and a gain request; an execution layer projects the gain request into the supplied set before execution. In a representative oblique insertion task, CG-RL achieves an overall constraint-compliant success rate of $85.8\pm7.7\%$ across five training seeds, compared with $50.1\pm1.0\%$ for the fixed midpoint controller. It adjusts its advance rate continuously with the prescribed force limit and generalizes to more permissive force limits above the training range. The force limit and gain set therefore serve different roles: the force limit conditions the policy, whereas the gain set constrains the controller.

Future work will pursue two directions. First, we will evaluate CG-RL using real construction materials and calibrate suitable force limits $F_{\max}$ from their measured contact responses. These results can be organized into a material database for selecting task-specific force limits for different materials. Second, we will extend CG-RL to insertion tasks with deformable
components, such as compliant pegs. The observation and constraint models will account for changes in contact geometry and force caused by deformation.

\bibliographystyle{IEEEtran}
\bibliography{references}

\end{document}